\documentclass[10pt,twocolumn,letterpaper]{article}

\usepackage{cvpr}
\usepackage{times}
\usepackage{epsfig}
\usepackage{graphicx}
\usepackage{amsmath}
\usepackage{amssymb}
\usepackage{amsfonts}
\usepackage{enumitem}
\usepackage{multirow}

\usepackage[linesnumbered,ruled]{algorithm2e}

\usepackage[pagebackref=true,breaklinks=true,letterpaper=true,colorlinks,bookmarks=false]{hyperref}

\cvprfinalcopy 

\def\cvprPaperID{3257} 

\ifcvprfinal\fi
\begin{document}

\title{Unsupervised Histopathology Image Synthesis}

\author{
	   Le Hou$^1$, Ayush Agarwal$^2$, Dimitris Samaras$^1$, Tahsin M. Kurc$^{3,4}$,
       Rajarsi R. Gupta$^{3,5}$, Joel H. Saltz$^{3,1,5,6}$
       \\
       $^1$Dept. of Computer Science, Stony Brook University\\
       $^2$Dougherty Valley High School, California\\
       $^3$Dept. of Biomedical Informatics, Stony Brook University\\
       $^4$Oak Ridge National Laboratory\\
       $^5$Dept. of Pathology, Stony Brook Hospital\\
       $^6$Cancer Center, Stony Brook Hospital\\
       {\tt\small\{lehhou,samaras\}@cs.stonybrook.edu} \qquad {\tt\small ayush94582@gmail.com}\\
       {\tt\small\{tahsin.kurc,joel.saltz\}@stonybrook.edu} \qquad {\tt\small rajarsi.gupta@stonybrookmedicine.edu}\\
}

\maketitle

\begin{abstract}
Hematoxylin and Eosin stained histopathology image analysis is essential for the diagnosis and study of complicated diseases such as cancer. Existing state-of-the-art approaches demand extensive amount of supervised training data from trained pathologists. In this work we synthesize in an unsupervised manner, large histopathology image datasets, suitable for supervised training tasks. We propose a unified pipeline that: a) generates a set of initial synthetic histopathology images with paired information about the nuclei such as segmentation masks; b) refines the initial synthetic images through a Generative Adversarial Network (GAN) to reference styles; c) trains a task-specific CNN and boosts the performance of the task-specific CNN with on-the-fly generated adversarial examples. Our main contribution is that the synthetic images are not only realistic, but also representative (in reference styles) and relatively challenging for training task-specific CNNs. We test our method for nucleus segmentation using images from four cancer types. When no supervised data exists for a cancer type, our method without supervision cost significantly outperforms supervised methods which perform across-cancer generalization. Even when supervised data exists for all cancer types, our approach without supervision cost performs better than supervised methods.
\end{abstract}

\section{Introduction}

\begin{figure}[t]
\begin{center}
   \includegraphics[width=0.99\linewidth]{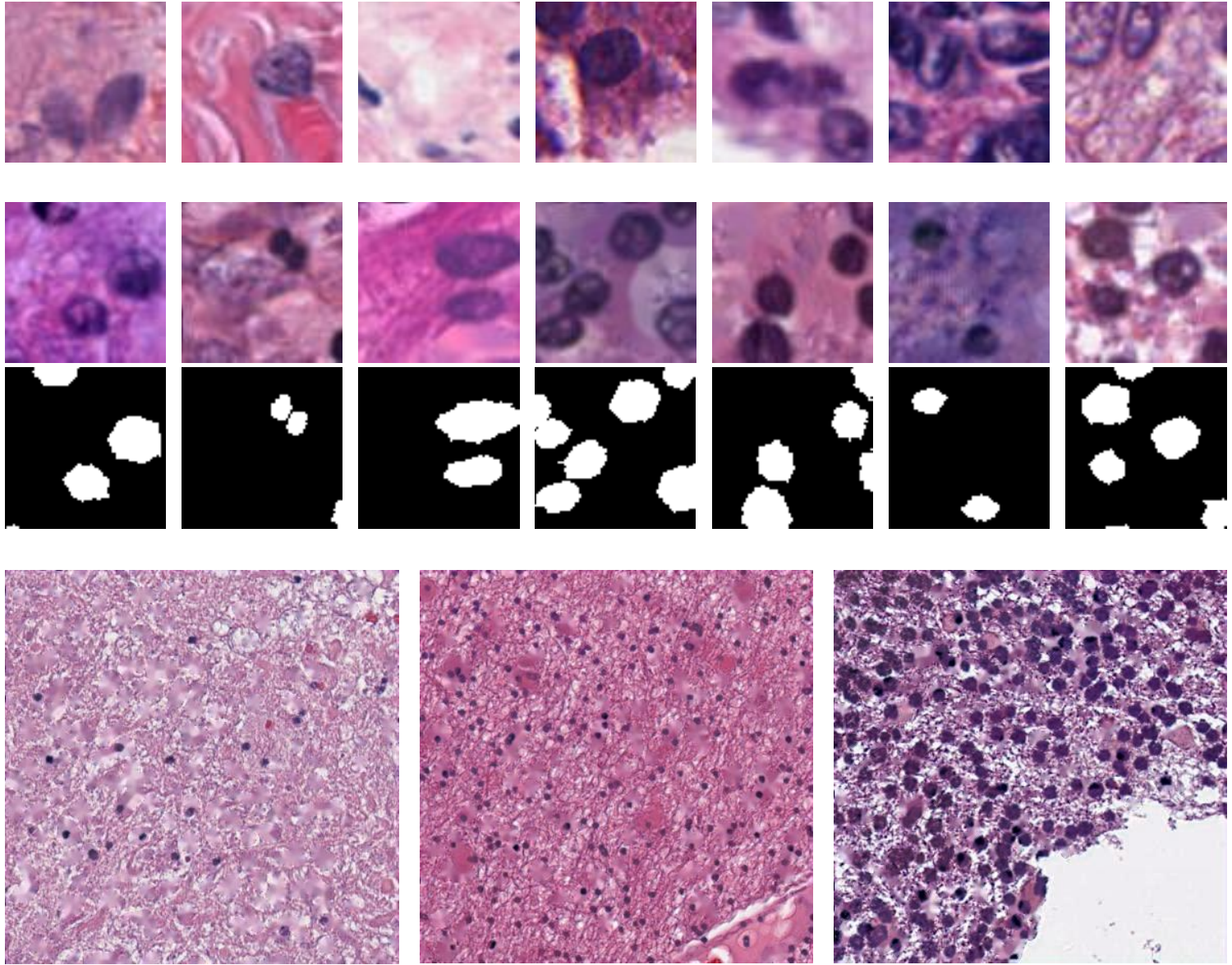}
\end{center}
   \caption{First row: real histopathology image patches at 40X magnification, with unknown nucleus segmentation mask. Center two rows: our synthesized histopathology image patches at 40X and corresponding nucleus segmentation masks. Last row: our synthesized 20X large patches with different cellularity and nuclear pleomorphism.}
\label{fig:first_figure}
\end{figure}

We propose a method for the synthesis of large scale, realistic image datasets that can be used to train machine learning algorithms for histopathology image analysis in precision medicine. Precision medicine requires the ability to classify patients into specialized cohorts that differ in 
their susceptibility to a particular disease, in the biology and/or prognosis of the disease, or in 
their response to therapy~\cite{national2011toward,collins2015new}. Imaging data and in particular 
quantitative features extracted by image analysis have been identified as a critical source of information 
particularly for cohort classification (imaging phenotypes) and tracking response to therapy. Quantitative 
features extracted from Pathology and Radiology imaging studies, provide valuable diagnostic and prognostic 
indicators of cancer~\cite{cooper2010integrative,cooper2012integrated,aerts2014decoding,parmar2015radiomic,gillies2015radiomics}.

\begin{figure*}[t]
\begin{center}
   \includegraphics[width=0.80\linewidth]{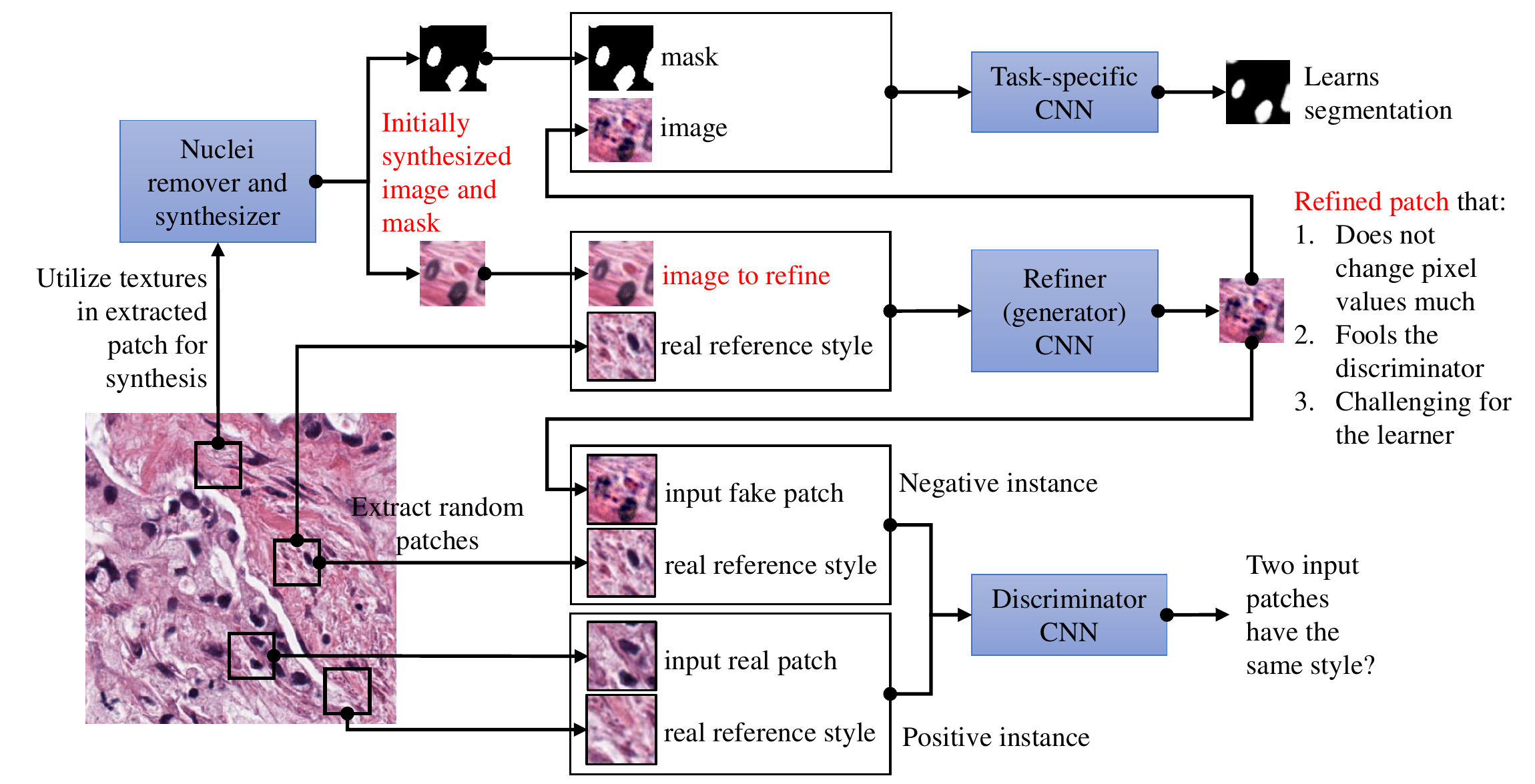}
\end{center}
   \caption{Our method synthesizes histopathology images with desired styles and known information (such as nuclei and their segmentation masks). There are three CNNs in our approach. The refiner (generator) CNN refines initial synthetic image patches synthesized by the ``nuclei-remover-and-synthesizer" module according to reference styles. The discriminator learns to criticize the refined patches, so that the refiner can generate realistic patches that match the reference style. The
   task-specific CNN learns to segment or classify the refined patches and give feedback to the refiner, so that the refiner can generate challenging patches for training. We show details of the ``nuclei-remover-and-synthesizer" in Fig.~\ref{fig:synthesize_pipeline}.}
\label{fig:method_overview}
\end{figure*}

Nucleus segmentation in histopathology images is a central component in virtually 
all Pathology precision medicine imaging studies~\cite{colen2014nci,gurcan2013digital,cooper2012digital,saltz2017towards}. 
Existing machine-learning based image analysis   methods~\cite{bayramoglu2016transfer,xu2016stacked,wang2016subtype,xie2015beyond,chen2017dcan,zhang2017deep,yang2017suggestive,hou2016patch,murthy2017center} largely rely on availability of 
large annotated training datasets.
One of the challenges is the 
generation of training datasets, because it requires the involvement of expert 
pathologists. It is a time-consuming, labor intensive and expensive process.
In our experience, manually segmenting a nucleus in a tissue image takes about 2 minutes. A relatively small 
training dataset of 50 representative 600$\times$600-pixel image patches has about 7000 nuclei.
This corresponds to 225 hours of a Pathologist's time to generate the training dataset.
Such a training dataset is still a very small sampling of a moderate size dataset of a few hundred
images -- each whole slide tissue image
has a few hundred thousand to over a million nuclei.
Moreover, the training phase usually should be repeated for different cancer types or even 
within a cancer type when new images are added. This is because of the heterogeneity of tissue specimens 
(of different cancer types, sub-types and stages) as well as variations arising from tissue 
preparation and image acquisition.

We propose a methodology to 
significantly reduce the cost of generating training datasets by synthesizing histopathology images that can be 
used for training task specific algorithms.
With our methodology a pathologist would only need to help tune the hyperparameters of the unsupervised synthesis pipeline by giving rounds of feedback (synthetic nuclei should be 20\% larger, \etc.). In this way the time cost of human involvement in training dataset generation would go down from hundreds of hours to under one hour.
In our experiments, we synthesized a dataset 400 times larger than a manually collected training set, which would cost 225 hours of a Pathologist's time. Due to the large volume of training data, segmentation CNNs trained on the synthetic dataset outperform segmentation CNNs trained on the more precise but much smaller manually collected dataset. 


Recent works in machine learning for image analysis have proposed crowd-sourcing or high-level, less accurate annotations, such as scribbles, to generate large training datasets by humans~\cite{lin2016scribblesup,vicente2016large,yang2017suggestive}. Another approach is to automatically synthesize training data, including pathology images and associated structures such as nucleus segmentation masks. Work by Zhou {\em et al.}~\cite{zhou2017evaluation} segments nuclei inside a tissue image and redistributes the segmented nuclei inside the image. The segmentation masks of the redistributed nuclei are assumed to be the predicted segmentation masks. Generative Adversarial Network (GAN)~\cite{radford2015unsupervised} approaches have been proposed for generation of realistic images~\cite{costa2017towards,bi2017synthesis,bayramoglu2017towards,shrivastava2017learning,calimeri2017biomedical,zhao2017dual,osokin2017gans}. 
For example, an image-to-image translation GAN~\cite{isola2017image,costa2017towards} 
synthesizes eye fundus images. However, it requires an accurate supervised segmentation network to segment eye vessels out, as part of the synthesis pipeline. The S+U learning framework~\cite{shrivastava2017learning} uses physics-based rendering methods to obtain initially synthesized images and refines via a GAN those images to increase their realism. This method achieves state-of-the-art 
results in eye gaze and hand pose estimation tasks.

There are several challenges to synthesizing histopathology images. 
First, state-of-the-art image synthesis approaches~\cite{shrivastava2017learning,zhao2017dual,richter2016playing,ronneberger2015u} require a physics-based 3D construction and rendering model. However, physics in the cellular level is largely unknown, making physics-based modeling infeasible. Second, histopathology images are heterogeneous with rich structure and texture characteristics. It is hard to synthesize images with a large variety of visual features. Moreover, care must be taken to avoid synthesizing images which can easily become biased and easy to classify, despite being realistic and heterogeneous. Our methodology (Fig.~\ref{fig:method_overview}) addresses these problems for Hematoxylin and Eosin (H\&E) stained histopathology images. H\&E is the mostly commonly used staining system for disease diagnosis and prognosis.

\begin{figure}[t]
\begin{center}
   \includegraphics[width=0.99\linewidth]{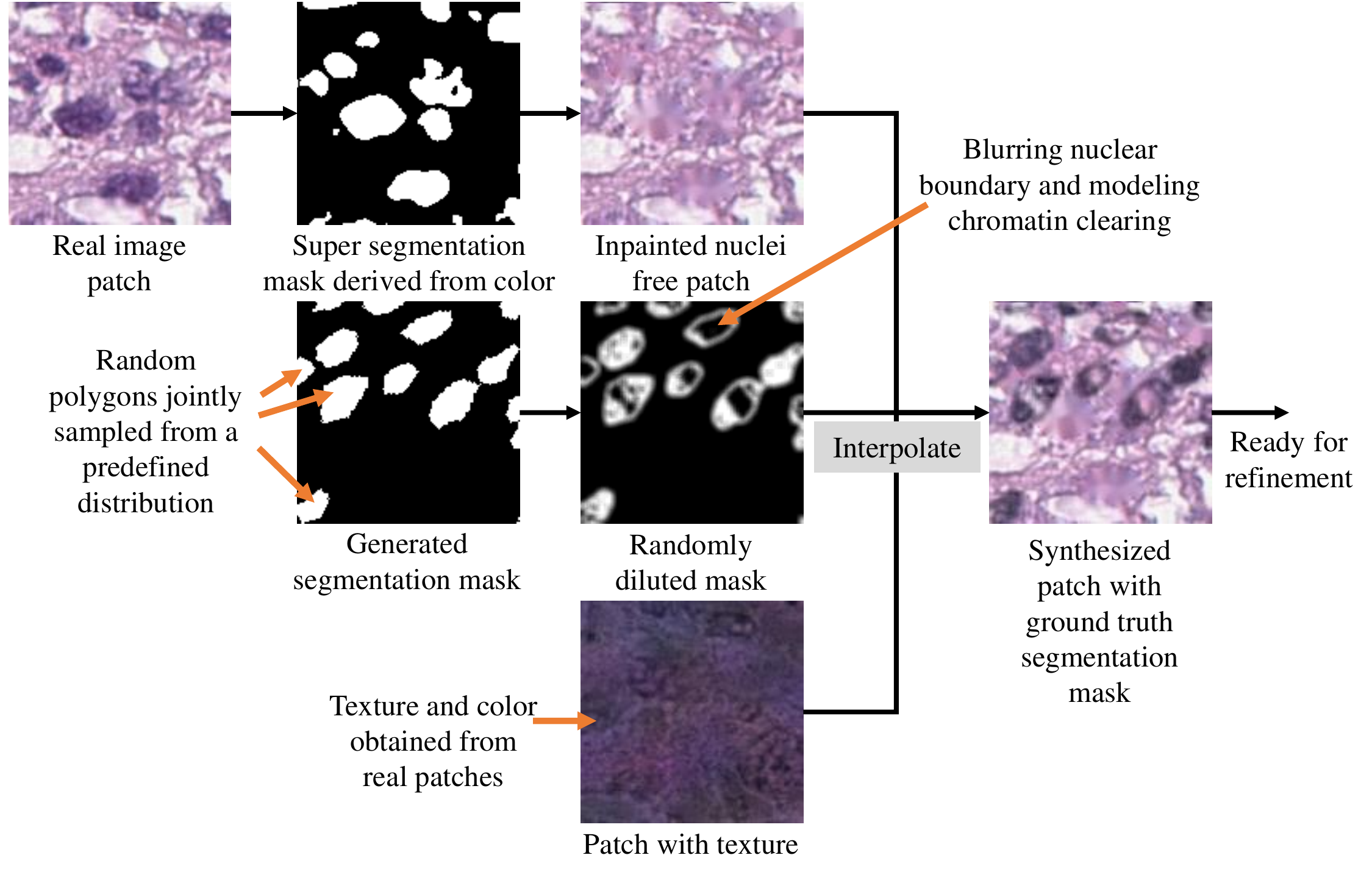}
\end{center}
   \caption{Inside the ``nuclei-remover-and-synthesizer" module: the process of synthesizing a histopathology image patch and nucleus segmentation mask in the initial stage. The synthesized image will be refined with GAN.}
\label{fig:synthesize_pipeline}
\end{figure}

The \textbf{first contribution} is a computer vision-based histopathology image synthesis method that generates initial synthetic histopathology images with desired characteristics such as the locations and sizes of the nuclei, cellularity, and nuclear pleomorphism, as shown in Fig.~\ref{fig:synthesize_pipeline}. Our method only needs a simple unsupervised segmentation algorithm that always super-segments nuclei. In ``super-segmentation", the segmented regions always fully contain the segmentation object.

The \textbf{second contribution} is that our method can synthesize heterogeneous histopathology 
images that span a variety of styles, i.e., tissue types and cancer subtypes.
Image synthesis methods essentially model the distribution of real data~\cite{li2017triple}. The joint distribution of real pixel values is very complex and hard to model. We propose to sample images from the real distribution and synthesizes images similar to the sampled real images, thus, simulating the distribution of real samples. Our model takes real images as references and generates realistic images in the reference style using a Generative Adversarial Network (GAN).
This can be viewed as an instance of universal style transfer~\cite{li2017universal,taigman2016unsupervised}.

Our \textbf{third contribution} is to train a task-specific model jointly with the image synthesis model. The image synthesis model is aware of the task-specific model and generates adversarial (hard) examples accordingly. Compared with existing hard example mining methods~\cite{shrivastava2016training,lemley2017smart} and adversarial data augmentation methods~\cite{goodfellow2014explaining}, our approach generates different versions of hard or adversarial training examples on-the-fly, according to the snapshot of the current task-specific model, instead of mining for existing hard examples in a dataset or inefficiently adding adversarial noise via slow optimization processes.

We test our method for nucleus segmentation using images from four cancer types. When no supervised data exists for a cancer type, our method without supervision cost significantly outperforms supervised methods which perform across-cancer generalization. Even when supervised data exists for all cancer types, our approach performed better than supervised methods.


\section{Initial Synthesis} \label{sec:initial-synthesis}
We utilize the texture characteristics of real histopathology image patches 
to generate initial synthetic images patches, in a background/foreground manner, with nuclei as the foreground.
The first step of this workflow is to create a synthetic image patch without any nuclei. The second step 
simulates the texture and intensity characteristics of nuclei in the real image patch. 
The last step combines the output from the first two steps based on a randomly generated 
nucleus segmentation mask (see Figure~\ref{fig:synthesize_pipeline} for the initial 
synthesized image patch). For simplicity, we will refer to image patches as images 
in the rest of the manuscript. Synthesizing a 200$\times$200 pixel patch at 40X magnification takes one second by a single thread on a desktop CPU.

\subsection{Generating Background Patches} \label{sec:generating-nuclei-free}
We first remove the foreground (nuclei) in an image patch to create a background image on which we will add synthetic nuclei. 
We apply a simple threshold-based super-segmentation method on the source image patch to 
determine nuclear pixels in the source image. In ``super-segmentation", the segmented regions always fully contain the segmentation object.
We then remove those pixels and replace them with 
color and texture values similar to the background pixels via image inpainting~\cite{telea2004image}.
Super-segmentation may not precisely 
delineate object boundaries and may include non-nuclear material in segmented 
nuclei. This is acceptable, because the objective of this step is to guarantee that only background tissue texture and intensity properties 
are used to synthesize the background image.

Hematoxylin mainly stains nucleic acids whereas Eosin stains proteins nonspecifically in tissue specimens~\cite{fischer2008hematoxylin}. We apply color deconvolution~\cite{ruifrok2001quantification} to H\&E images to obtain the Hematoxylin, Eosin, DAB (HED) color space. We threshold the H channel for nuclei segmentation. Specifically, we first decide the percentage of nuclear pixels, $p$, based on the average color intensity $h$, of th H channel. For $h$ in ranges $(-\infty,-1.25)$, $[-1.25,-1.20)$, $[-1.20,-1.15)$, $[-1.15,-1.10)$, $[-1.10,\infty)$, we set the percentage of nuclear pixels $p$ as 15\%, 20\%, 25\%, 30\%, 35\% respectively. These hyperparameters were selected by visually inspecting super-segmentation results on a set of image patches from all cancer types in the TCGA repository~\cite{TCGAdataset}. The segmentation threshold, $t$, is the $p$-th percentile value of the H channel. After thresholding the H channel with $t$, we apply Gaussian smoothing to remove noise such as very small segmented regions. Finally, the segmented pixels are inpainted in a computationally efficient manner~\cite{telea2004image}.

\begin{figure}[t]
\begin{center}
   \includegraphics[width=0.99\linewidth]{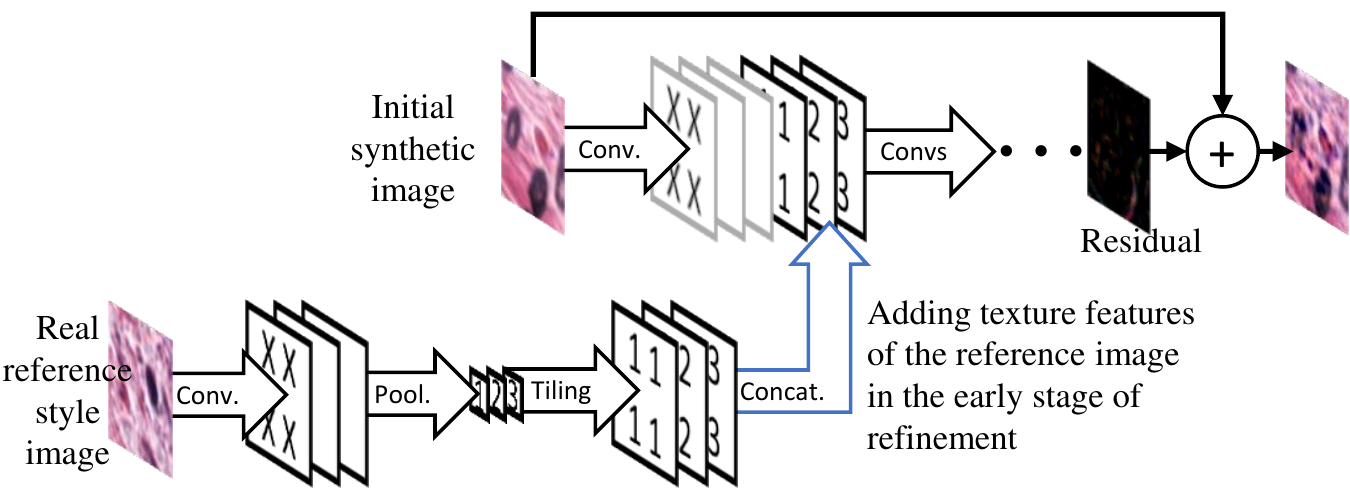}
\end{center}
   \caption{Our refiner (generator) CNN adds the global texture features of the reference image into the early stage of refinement, so that the initial synthetic image will be refined according to the textures of the reference style image.}
\label{fig:refiner_cnn}
\end{figure}

\subsection{Simulating Foreground Textures}
One approach to simulating foreground nuclear textures is to apply a sub-segmentation method and gather nuclear textures from segmented regions. In ``sub-segmentation", the segmentation object always contains segmented regions. The objective of sub-segmentation ensures that pixels within the nuclei are used for nuclei synthesis. Since nuclei are generally small and make up a small portion of the tissue area-wise, sub-segmentation will yield very limited amount of nuclear material which is not enough for existing reconstruction methods to generate realistic nuclear material patches. Thus, our approach utilizes textures in the Eosin channel~\cite{fischer2008hematoxylin} of a randomly extracted real patch (different from the background source patch in Section~\ref{sec:generating-nuclei-free}) and combines them with nuclear color obtained via sub-segmentation of the input patch to generate nuclear textures.

We have observed that this method gives realistic textures. To sub-segment, we use the same process as for the super-segmentation approach but with different $p$ values: For $h$ in ranges $(-\infty,-1.25)$, $[-1.25,-1.20)$, $[-1.20,-1.15)$, $[-1.15,-1.10)$, $[-1.10,\infty)$, we set $p$ as 10\%, 16\%, 21\%, 27\%, 32\% respectively.

\subsection{Combining Foreground and Background}
We  generate a nuclear mask and combine nuclear and non-nuclear textures according to the mask. First, we randomly generate non-overlapping polygons with variable sizes and irregularities. To model the correlation between the shapes of nearby nuclei, we distort all polygons by a random quadrilateral transform. The resulting nucleus mask is regarded as a synthetic ``ground truth" segmentation mask. We then combine foreground and background patches by: 
\begin{equation}
I_{i,j} = A_{i,j} M_{i,j} + B_{i,j} (1-M_{i,j}) \text{.}
\label{eq:initial-syn-combine}
\end{equation}
Here, $I_{i,j}$ is the pixel value of the resulting synthetic image. Pixel values at position $i,j$ in the nuclear texture patch, in the nucleus free patch, and in the nucleus mask are denoted as $A_{i,j}$, $B_{i,j}$, $M_{i,j}$ respectively. 

Applying Eq.~\ref{eq:initial-syn-combine} naively results in significant artifacts, such as obvious nuclear boundaries. Additionally, clearing of chromatin cannot be modeled. To remedy these issues, we randomly clear the interior and blur the  boundaries of the polygons in $M$, before applying Eq.~\ref{eq:initial-syn-combine}.

\section{Refined Synthesis}
\label{sec:second-level-refinement}
We refine the initial synthetic images via adversarial training as shown in Fig.~\ref{fig:method_overview}. This phase implements a Generative Adversarial Network (GAN) model and consists of a refiner (generator) CNN and a discriminator CNN.

Given an input image $I$ and a reference image $S$, the refiner $\mathrm{G}$ with trainable parameters $\theta_G$ outputs a refined image $\tilde{I}=\mathrm{G}(I, S; \theta_G)$. Ideally, the output image is:
\begin{description}[leftmargin=0.05in]
\setlength\itemsep{0.1em}
\item[Regularized] The pixel-wise difference between the initial synthetic image and the refined 
image is small enough so that the synthetic ``ground truth'' remains unchanged.
\item[Realistic] It has a realistic representation of the style of the reference image.
\item[Informative/hard] It is a challenging case for the task-specific CNN so that the 
trained task-specific CNN will be robust.
\end{description}

We build three losses: $L_G^\mathrm{reg}$, $L_G^\mathrm{real}$, $L_G^\mathrm{hard}$, for each of the properties above. The weighted average of these losses as the final loss $L_G$ for training of the refiner CNN is:
\begin{equation}
\label{eq:refiner_final_loss}
L_G = \alpha L_G^\mathrm{reg} +\beta L_G^\mathrm{real} +\gamma L_G^\mathrm{hard} \text{.}
\end{equation}
Selection of hyperparameters $\alpha$, $\beta$, $\gamma$ is described in Sec.~\ref{sec:experiments}.

The regularization loss $L_G^\mathrm{reg}$ is defined as:
\begin{equation}
\label{eq:reg_loss}
L_G^\mathrm{reg}(\theta_G)=\mathrm{E}\big[\lambda_1 ||I - \tilde{I}||_1 + \lambda_2 ||I - \tilde{I}||_2\big] \text{,}
\end{equation}
where $\mathrm{E}[\cdot]$ is the expectation function applied on the training set, $||I - \tilde{I}||_1$ and $||I - \tilde{I}||_2$ are the $L$-1 and $L$-2 norms of $I - \tilde{I}$ respectively and $\lambda_1$ and $\lambda_2$ are predefined parameters. This is the formulation of second order elastic net regularization~\cite{zou2005regularization}. In practice, we select the lowest $\lambda_1$ and $\lambda_2$ possible that do not result in significant visual changes of $\tilde{I}$ compared to $I$. 

The loss for achieving a realistic reference style is:
\begin{equation}
\label{eq:realism_loss}
L_G^\mathrm{real}(\theta_G) = \mathrm{E}\big[\mathrm{log}\big(1-\mathrm{D}(\tilde{I}, S; \theta_D)\big)\big] \text{,}
\end{equation}
where $\mathrm{D}(\tilde{I}, S; \theta_D)$, is the output of the discriminator $\mathrm{D}$ with trainable parameters $\theta_D$ given the refined image $\tilde{I}$ and the same reference style image $S$ as input. It is the estimated probability by $\mathrm{D}$ that input $\tilde{I}$ and $S$ are real images in the same style.

\begin{figure*}[t]
\begin{center}
   \includegraphics[width=0.9\linewidth]{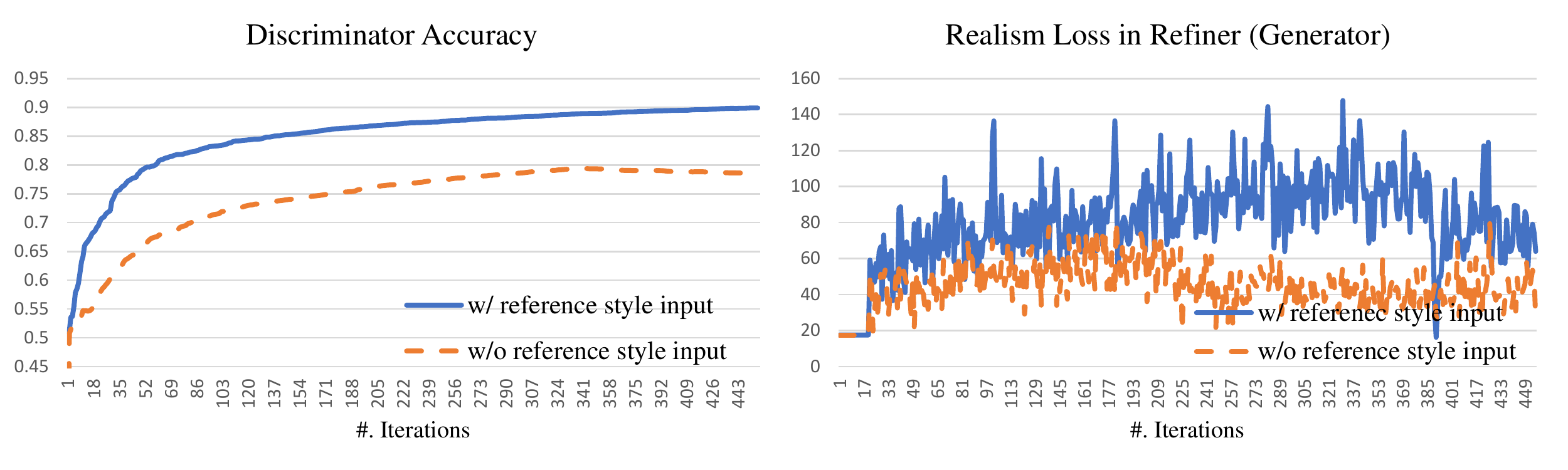}
\end{center}
   \caption{The effect of introducing real reference style images in the GAN training process. To fool the discriminator that ``knows'' the reference style, the refined images should be in the same style as the reference image, in addition to being realistic. Thus, the discriminator with reference style input is more accurate, and gives significantly more feedback in terms of the realism loss (Eq.~\ref{eq:realism_loss}) to the refiner.}
\label{fig:effect_of_reference_discrim}
\end{figure*}

The Discriminator $D$ with trainable parameters $\theta_D$ has two types of input: pairs of real images within the same style $\langle S', S\rangle$ and a pair with one synthetic image $\langle \tilde{I}, S\rangle$. The loss of $D$ is defined as:
\begin{multline}
\label{eq:discrim_loss}
L_D(\theta_D) = -\mathrm{E}\big[\log \big(\mathrm{D}(S', S; \theta_D)\big)\big] \\
- \mathrm{E}\big[\log\big(1-\mathrm{D}(\tilde{I}, S; \theta_D)\big)\big] \text{.}
\end{multline}
The discriminator learns to maximize its output probability for real pairs $\langle S', S\rangle$ and minimize it for $\langle \tilde{I}, S\rangle$. By introducing the reference style image $S$, the discriminator can correctly recognize the pair that contains a synthetic image if the synthetic image is not realistic, or it has a different style compared to the reference style image.

\paragraph{CNN Architecture for Style Transfer}
The generator and discriminator both take a reference image and refine or classify the other input image according to textures in the reference image. We implement this feature with a CNN which takes two input images. Existing CNN architectures, such as the siamese network~\cite{chopra2005learning,koch2015siamese}, merge or compare the features of two input images at a late network stage. However, the generator must represent the textures in the reference image and use it in the process of refinement at a early stage. To achieve this, our network has two branches: the texture representation branch and the image refinement branch. As is shown in Fig.~\ref{fig:refiner_cnn}, the texture representation branch takes the reference image as input and outputs a feature vector representing the reference image. The image refinement branch takes both the initial synthetic image and the reference image and generates 
a refined image.

We show the effect of adding the reference style images in GAN training in Fig.~\ref{fig:effect_of_reference_discrim}. The discriminator is significantly more accurate and gives more feedback in terms of the realism loss $L_G^\mathrm{real}(\theta_G)$, to the refiner.

\section{On-the-fly Hard Example Synthesis}
\label{sec:on-the-fly-hard}
The refiner is trained with loss $L_G^\mathrm{hard}$ to generate challenging training examples (with larger loss) for the task-specific CNN. We simply define $L_G^\mathrm{hard}$ as the negative of the task-specific loss:
\begin{equation}
\label{eq:learner_adv_loss}
L_G^\mathrm{hard}(\theta_G) = -L_R(\theta_R) \text{,}
\end{equation}
where $L_R(\theta_R)$ is the loss of a task-specific model $R$ with trainable parameters $\theta_R$. In the case of segmentation, $L_R(\theta_R)$ is the conventional segmentation loss used in deep learning~\cite{long2015fully,noh2015learning}. When training the refiner, we update $\theta_G$ to produce refined images that maximizes $L_R$. When training the task-specific CNN, we update $\theta_R$ to minimize $L_R$.

The underlying segmentation ground truth of the refined images would change significantly if $L_G^\mathrm{hard}(\theta_G)$ overpowered $L_G^\mathrm{reg}(\theta_G)$. We down weight $L_G^\mathrm{hard}$ by a factor of $0.0001$ to minimize the likelihood of this outcome.

\paragraph{Training process} We randomly initialize the refiner, discriminator and the task-specific networks. During the training process, the realism loss $L_G^\mathrm{real}$ and the task-specific adversarial loss $L_G^\mathrm{hard}$ are fed back to the refiner from the discriminator and the task-specific CNNs respectively. However, because we randomly initialize the discriminator and the task-specific networks, these feedbacks are initially useless for the refiner. Following the existing image refining GAN~\cite{shrivastava2017learning}, we initially train each CNN individually before training them jointly. The process is summarized in Alg.~\ref{alg:training_process}.

\begin{algorithm}
\label{alg:training_process}
\SetKwInOut{Input}{Input}
\SetKwInOut{Output}{Output}
\Input{A set of training images. Number of training iterations $N_G$, $N_D$, $N_R$, $N_{GD}$, $N_{GDR}$, $n_G$, $n_D$, $n_R$. Loss parameters $\alpha$, $\beta$, $\gamma$, $\lambda_1$, $\lambda_2$.
}
\Output{Trained segmentation/classification CNN $R$.}
Randomly initialize the trainable parameters $\theta_G$, $\theta_D$ and $\theta_R$ in $G$, $D$ and $R$ respectively.

Train $G$ to minimize $L_G^\mathrm{reg}(\theta_G)$ for $N_G$ iterations.

Train $D$ to minimize $L_D(\theta_D)$ for $N_D$ iterations.

\For{$n = 1,\dots,N_{GD}$}
{
Train $G$ to minimize $\alpha L_G^\mathrm{reg}(\theta_G) + \beta L_G^\mathrm{real}(\theta_G)$ for $n_G$ iterations.

Train $D$ to minimize $L_D(\theta_D)$ for $n_D$ iterations.
}

Train $R$ to minimize $L_R(\theta_R)$ for $N_R$ iterations.

\For{$n = 1,\dots,N_{GDR}$}
{
Train $G$ to minimize $\alpha L_G^\mathrm{reg}(\theta_G) + \beta L_G^\mathrm{real}(\theta_G) + \gamma L_G^\mathrm{hard}(\theta_G)$ for $n_G$ iterations.

Train $D$ to minimize $L_D(\theta_D)$ for $n_D$ iterations.

Train $R$ to minimize $L_R(\theta_R)$ for $n_R$ iterations.
}

return $R$ with $\theta_R$;
\caption{Refining and task-specific learning.}
\end{algorithm}

\begin{figure*}[t]
\begin{center}
   \includegraphics[width=0.95\linewidth]{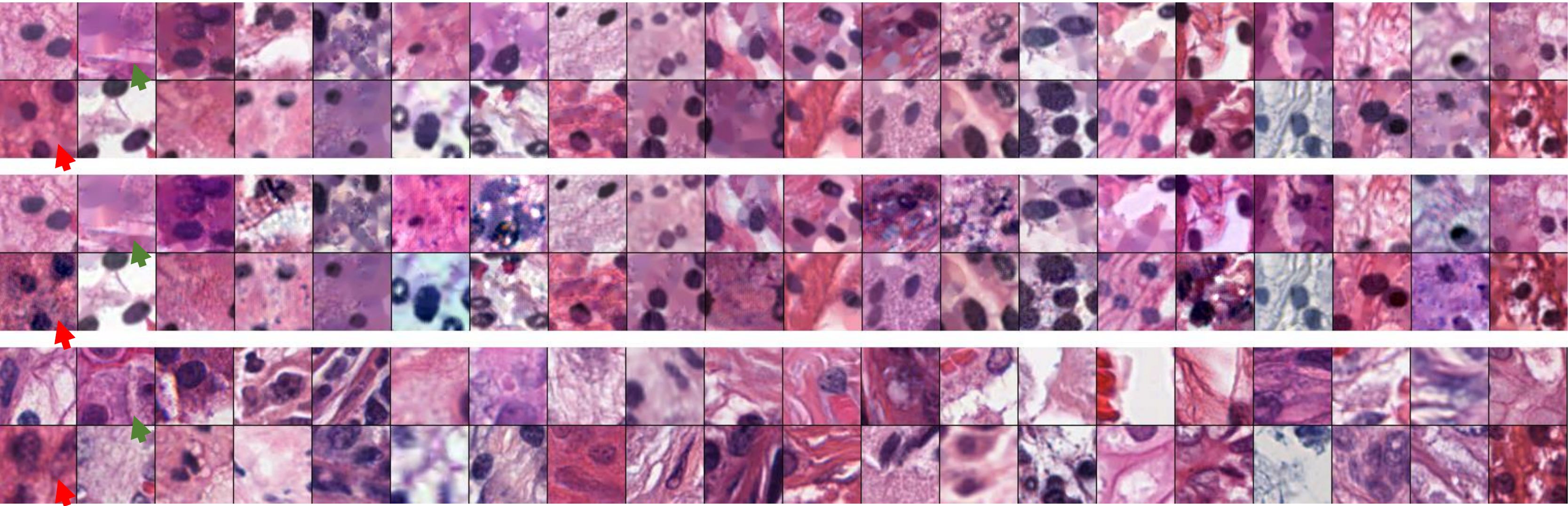}
\end{center}
   \caption{Randomly selected examples of initial synthetic histopathology images (first two rows), refined images (second two rows), and corresponding real reference style images (last two rows). The refiner successfully refines the initial synthetic images to reference styles without modifying the images significantly (example indicated by red arrow). On cases where the refiner fails, this signifies that the initial synthetic images can not be transfered to reference styles without significantly modifying the images (sample indicated by green arrow).}
\label{fig:nuclei_samples}
\end{figure*}

\begin{figure}[t]
\begin{center}
   \includegraphics[width=0.85\linewidth]{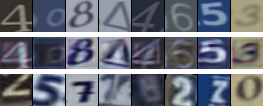}
\end{center}
   \caption{Randomly selected examples of initial synthetic street view house number images (first row), refined images (second row), and corresponding real reference style images (last row).}
\label{fig:SVHN_samples}
\end{figure}

\section{Visual Test by Expert}
\label{sec:visual-user-study}
\begin{figure}
\begin{center}
   \includegraphics[width=0.90\linewidth]{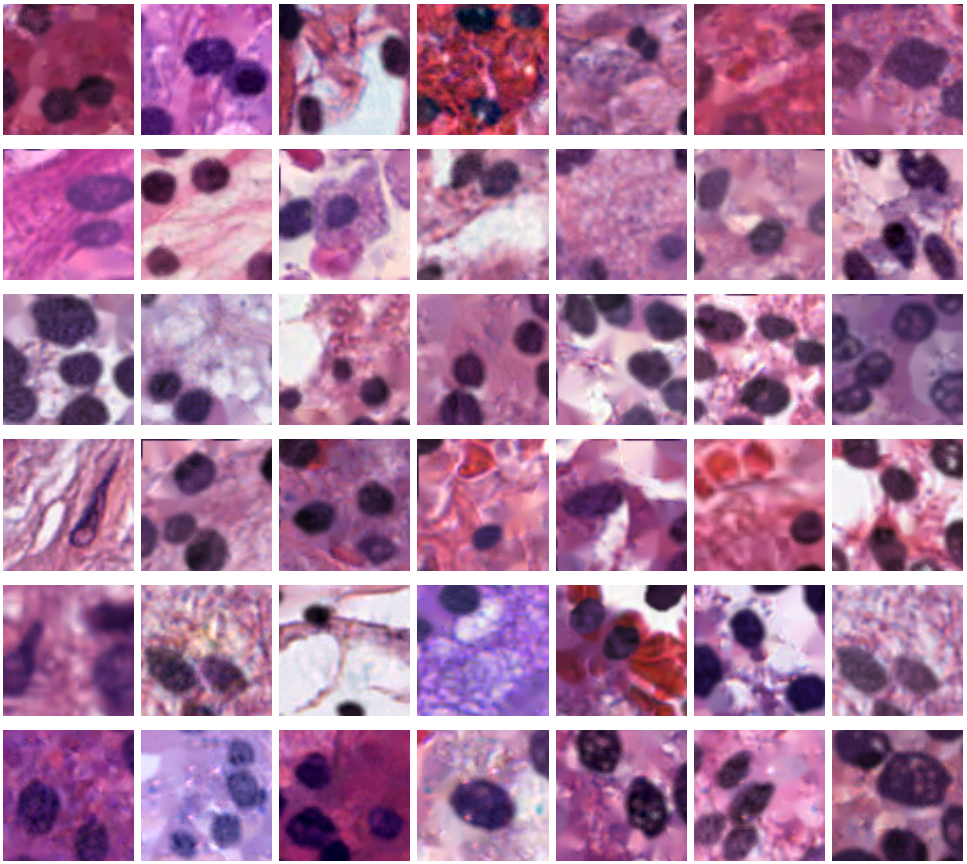}
\end{center}
   \caption{Representative examples of synthetic images that appeared real to the pathologist.}
\label{fig:examples}
\end{figure}

To verify that the synthetic images are realistic, we asked a pathologist to distinguish real versus synthetic images. In particular, we showed the pathologist 100 randomly extracted real patches, 100 randomly selected initial synthetic patches, and 100 randomly selected refined patches. Out of this set, the pathologist selected the patches he thought were real. We summarize the results in Table~\ref{tab:visual-user-study}. A significant number of initial synthetic images (46\%) were classified as real by the pathologist. Most of the refined patches (64\%) were classified real. Note that 17\% of the real patches were classified fake. This is because many of those image patches are out-of-focus or contain no nuclei. In average, the pathologist spend 4.6 seconds classifying on each patch. We show representative examples of synthetic images that appeared real to the pathologist in Fig.~\ref{fig:examples}.

\begin{table}[ht]
	\centering
	\begin{tabular}{r | p{1.27cm} | p{1.27cm} }
	\hline
    Ground truth & \#. classified real & \#. classified fake \\
    \hline
    Initial synthetic & 46 & 54 \\
    Refined & 64 & 36 \\
    Real & 87 & 13 \\
	\hline
	\end{tabular}
\vspace{0.1cm}
\caption{We show 100 randomly selected and ordered initial synthetic, refined and real patches to a pathologist, and ask the pathologist to classify them as real or fake.}
\label{tab:visual-user-study}
\end{table}

We show randomly selected examples of initial synthetic and refined histopathology images in Fig.~\ref{fig:nuclei_samples}. The refiner successfully refines the initial synthetic images to reference styles without modifying the images significantly. On cases where the refiner fails, the initial synthetic images can not be transfered to the reference styles without significantly modifying the images.

\begin{figure}
\begin{center}
   \includegraphics[width=0.85\linewidth]{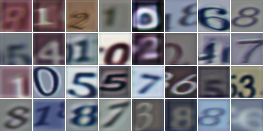}
\end{center}
   \caption{Randomly selected examples of refined synthetic street view house number images.}
\label{fig:examples_SVHN}
\end{figure}

To demonstrate the generality of our method, and how our method works outside the pathology domain, we synthesize house street numbers using the SVHN database~\cite{netzer2011reading}. To generate initial synthetic images from real images, we apply a k-means clustering method to obtain the background and foreground colors in the real images. Then we write a digit in a random font in constant foreground color. The refiner refines the style of the
initial synthetic images to the real reference style. We show randomly selected examples in Fig.~\ref{fig:SVHN_samples} and Fig.~\ref{fig:examples_SVHN}.

\section{Experiments} \label{sec:experiments}
To evaluate the performance of our method, we conducted experiments with ground-truth datasets generated for the MICCAI15 and MICCAI17 nucleus segmentation challenges~\cite{miccai2015segmentation,miccai2017segmentation}. 
Additionally, we synthesized large pathology image patches for two classes: high/low cellularity and nuclear pleomorphism and show that a task-specific CNN trained on this dataset can classify glioblastoma (GBM) versus low grade gliomas (LGGs).

\subsection{Implementation Details}
\label{sec:implementation_details}
The refiner network, outlined in Fig.~\ref{fig:refiner_cnn}, has 21 convolutional layers and 2 pooling layers. The discriminator network has the same overall architecture with the refiner. It has 15 convolutional layers and 3 pooling layers. As the task-specific CNN, we implement U-net~\cite{ronneberger2015u} and a network with 15 convolutional layers and 2 pooling layers, and a semi-supervised CNN~\cite{hou2017sparse} for segmentation. We use a 11 convolutional layer network for classification. For hyperparameters in Eq.~\ref{eq:refiner_final_loss} and Eq.~\ref{eq:reg_loss}, we select $\alpha=1.0$, $\beta=0.7$, $\gamma=0.0001$, $\lambda_1=0.001$, $\lambda_2=0.01$ by validating on part of a synthetic dataset. We implement our method using an open source implementation of S+U learning~\cite{SU_implementation,shrivastava2017learning}. The methods we test are listed below.
\begin{description}
\item[Synthesis CAE-CNN] Proposed method with the semi-supervised CNN~\cite{hou2017sparse} as the task-specific segmentation CNN.
\item[Synthesis U-net] Proposed method with U-net~\cite{ronneberger2015u} as the task-specific segmentation CNN.
\item[Synthesis CNN] Proposed method with a 15 layer segmentation network or a 11 layer classification network.
\item[CAE-CNN / U-net / CNN with supervision cost] We use the semi-supervised CNN~\cite{hou2017sparse}, U-net~\cite{ronneberger2015u} and the 15 layer CNN as standalone supervised networks, trained on real human annotated datasets. We augment the real images by rotating four times, mirroring, and rescaling six times.
\end{description}

\begin{figure*}[t]
\begin{center}
   \includegraphics[width=0.75\linewidth]{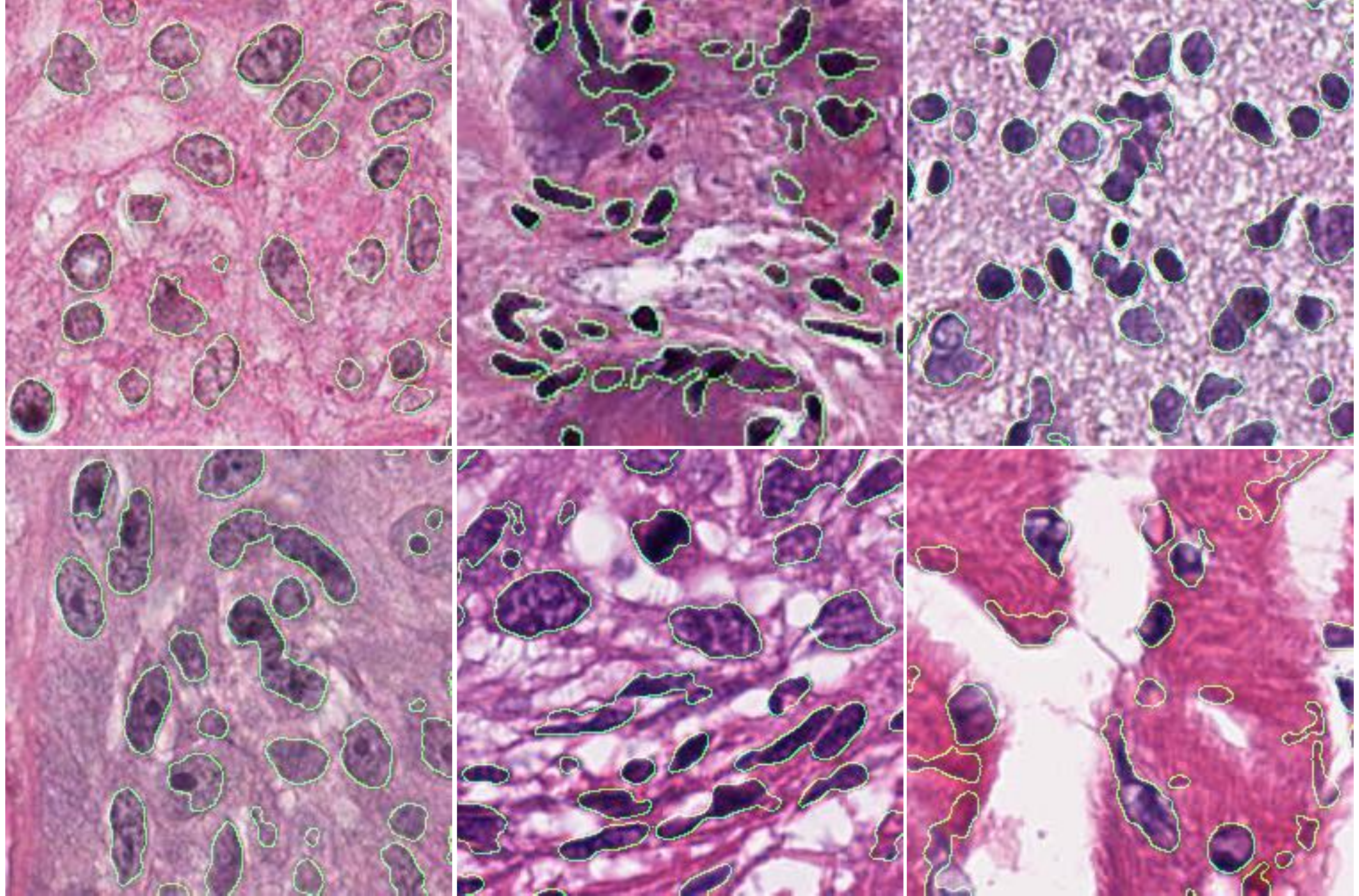}
\end{center}
   \caption{Randomly selected examples of nucleus segmentation results (green contours) on the MICCAI15 and MICCAI17 nucleus segmentation test set.}
\label{fig:segmentation_visual}
\end{figure*}

\begin{figure*}[t]
\begin{center}
   \includegraphics[width=0.99\linewidth]{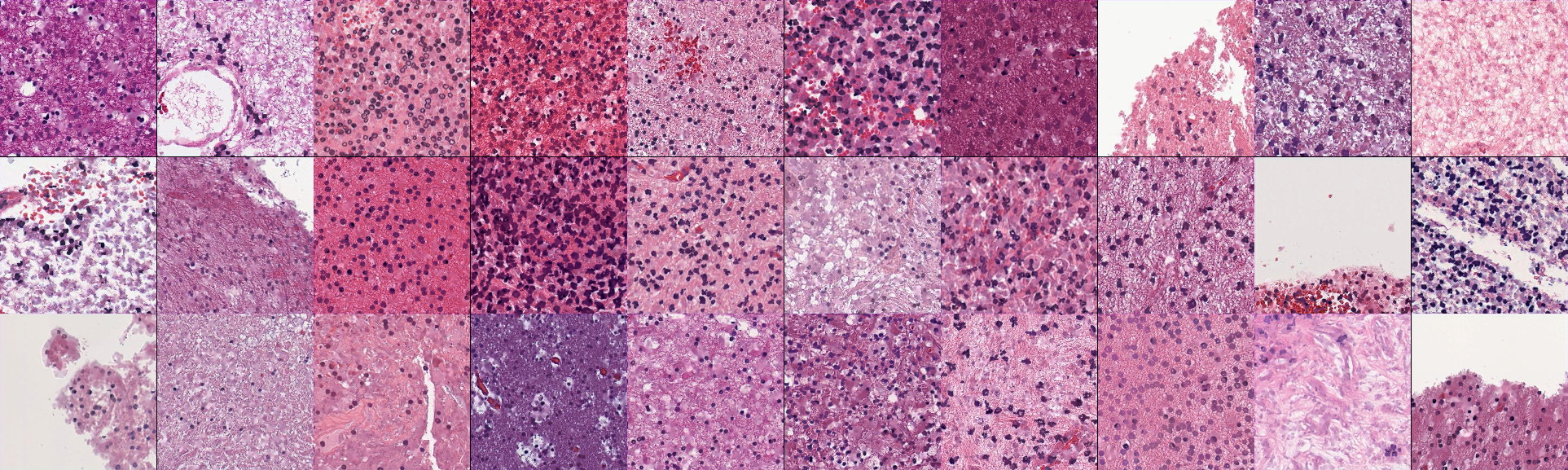}
\end{center}
   \caption{Randomly selected examples of synthetic 384$\times$384 pixel 20X histopathology image patches with various levels of cellularity and nuclear pleomorphism.}
\label{fig:20X_samples}
\end{figure*}

\subsection{Nucleus segmentation}
The MICCAI15 nucleus segmentation challenge dataset~\cite{miccai2015segmentation} contains 15 training and 18 testing images extracted from whole slide images of GBM and LGG. The MICCAI17 dataset~\cite{miccai2017segmentation} contains 32 training and 32 testing images, extracted from whole slide images of GBM, LGG, Head and Neck Squamous cell Carcinoma (HNSC) and Lung Squamous Cell Carcinoma (LUSC). A typical resolution is 600$\times$600 pixels at 20X or 40X (0.50 or 0.25 microns per pixel) magnifications. Assuming that annotating one nucleus takes 2 minutes, it would take about 225 man-hours to generate these training datasets. In contrast, it would take just one hour of a Pathologist for us to synthesize a large dataset.

We investigate if the task-specific supervised method performs better in standalone mode when it is trained on a few but real training data or when it is trained with abundant but synthetic training data generated by our synthesis pipeline. We evaluate the supervised segmentation method of Sec.~\ref{sec:implementation_details} under three scenarios:
\begin{description}
\item[Universal] We train one universal segmentation CNN on training images of all two/four (MICCAI15/MICCAI17) cancer types.
\item[Cancer specific] We train one CNN for each cancer type. During test time, we apply the corresponding CNN based on the cancer type of the input image.
\item[Across cancer] To evaluate the performance of supervised CNNs on cancer types that lack training data, we train one CNN for each cancer type in the testing set, excluding training images of that cancer type from the training set. During test time, based on the cancer type of the input image, we apply the corresponding CNN that was not trained with that cancer type.
\end{description}
Most cancer types do not have a significant nucleus segmentation training set. Therefore, the third scenario is a very common 
real world use case. For our method, we generated 200k 75$\times$75-pixel initial synthetic patches at 40X magnification for each cancer type.

\begin{table}[ht!]
	\centering
	\begin{tabular}{p{6.2cm} | p{1cm}}
	\hline
	 Segmentation methods & DICE Avg. \\
    \hline
     Synthesis CAE-CNN (proposed) & 0.8424 \\
     CAE-CNN with supervision cost, Universal & 0.8362 \\
    \hline
     Synthesis U-net (proposed) & 0.8063 \\
     U-net with supervision cost, Universal & 0.7984 \\
    \hline
     Synthesis CNN (proposed) & 0.8254 \\
     CNN with supervision cost, Universal & 0.8013 \\
     CNN with supervision cost, Cancer specific & 0.8032 \\
     CNN with supervision cost, Across cancer & 0.7818 \\
    \hline
     Supervised contour-aware net (challenge winner)~\cite{chen2017dcan} & 0.812 \\
	\hline
	\end{tabular}
\vspace{0.1cm}
\caption{Nucleus segmentation results on the MICCAI15 nucleus segmentation dataset. On cancer types without annotated training data, our approach outperforms the supervised method (CNN with supervision cost, Across cancer) significantly. Even when supervised data exists for all cancer types, our approach improves the state-of-the-art performance without any supervision cost.}
\label{tab:miccai15_results}
\end{table}

\begin{table}[ht!]
	\centering
	\begin{tabular}{l | l}
	\hline
	 Segmentation methods & DICE avg \\
    \hline
     Synthesis CAE-CNN (proposed) & 0.7731 \\
     CAE-CNN with supervision cost, Universal & 0.7681 \\
    \hline
     Synthesis U-net (proposed) & 0.7631 \\
     U-net with supervision cost, Universal & 0.7645 \\
    \hline
     Synthesis CNN (proposed) & 0.7738 \\
     CNN with supervision cost, Universal & 0.7713 \\
     CNN with supervision cost, Cancer specific & 0.7653 \\
     CNN with supervision cost, Across cancer & 0.7314 \\
    \hline
     Challenge winner & 0.783 \\
	\hline
	\end{tabular}
\vspace{0.1cm}
\caption{Nucleus segmentation results on the MICCAI17 nucleus segmentation dataset. On cancer types without annotated training data, our approach outperforms the supervised method (CNN with supervision cost, Across cancer) significantly. Even when supervised data exists for all cancer types, our approach matches the state-of-the-art performance without any supervision cost.}
\label{tab:miccai17_results}
\end{table}

We use the average of two versions of DICE coefficients. Quantitative evaluation results on the MICCAI15 and MICCAI17 segmentation datasets are shown in Tab.~\ref{tab:miccai15_results} and Tab.~\ref{tab:miccai17_results}. With cancer types without annotated training images, our approach outperforms the supervised method (CNN with supervision cost, Across cancer) significantly. Even when supervised data exists for all cancer types, our approach achieves state-of-the-art level performance or better without any supervision cost. We see that the supervised method we incorporated into our pipeline, has comparable performance to the winners of the two challenges.

\subsection{Ablation study}
We evaluate the importance of two proposed components of our method: utilizing a real reference style image for refinement 
and generating on-the-fly hard examples for CNN training. In particular, we remove one feature at a time and evaluate the performance of nucleus segmentation. Experimental results are shown in Tab.~\ref{tab:miccai17_ablation}. We see that both proposed methods improve the segmentation results. We also show the effect of introducing real reference style images as additional network inputs in Fig.~\ref{fig:effect_of_reference_discrim}. 

\begin{table}[ht!]
	\centering
	\begin{tabular}{l | l}
	\hline
	 Segmentation methods & DICE avg \\
    \hline
     Synthesis CNN (proposed) & 0.7738 \\
     No reference style during refinement & 0.7589 \\
     No on-the-fly hard examples & 0.7491 \\
	\hline
	\end{tabular}
\vspace{0.1cm}
\caption{Ablation study using the MICCAI17 nucleus segmentation challenge dataset. Each proposed method reduces the segmentation error by 6\% to 9\%.}
\label{tab:miccai17_ablation}
\end{table}

\subsection{Glioma classification}
We synthesize patches of 384$\times$384 pixels in 20X of two classes: relatively low cellularity and nuclear pleomorphism, versus relatively high cellularity and nuclear pleomorphism (Fig.~\ref{fig:20X_samples}). Cellularity and nuclear pleomorphism levels provide diagnostic information. We train the task-specific CNN to classify high versus low cellularity and nuclear pleomorphism patches. The cellularity and nuclear pleomorphism prediction results on real slides can distinguish Glioblastoma (GBM) versus Lower Grade Glioma (LGG) with an accuracy of 80.1\% (Chance being 51.3\%). A supervised approach~\cite{mousavi2015automated} trained for the GBM/LGG classification achieved an accuracy of 85\% using a domain specific pipeline with nucleus segmentation and counting.

\subsection{SVHN classification}
These experiments evaluate our method with the format1 sub-set in the Street View House Number (SVHN) dataset~\cite{netzer2011reading}. The subset contains 68,120 training images and 23549 testing images in 32$\times$32 pixels. We synthesized 68,120 images with digits and refined them to reference styles sampled in the format1 training set. Classification errors (1$-$accuracy) are shown in Tab.~\ref{tab:SVHN}.
\begin{table}[h]
	\centering
	\begin{tabular}{p{1.8cm} | r | r}
	\hline
	 Methods & Training set & Error \\
    \hline
     \multirow{5}{\hsize}{Synthesis CNN (proposed)} & 3,000 syn. training images & 29.03\% \\
      & 5,000 syn. training images & 23.24\% \\
      & 10,000 syn. training images & 18.47\% \\
      & 30,000 syn. training images & 17.57\% \\
      & 68,120 syn. training images & 17.08\% \\
    \hline
     \multirow{5}{\hsize}{CNN with supervision cost} & 3,000 \hspace{0.001cm} real training images & 24.55\% \\
      & 5,000 \hspace{0.001cm} real training images & 18.53\% \\
      & 10,000 \hspace{0.001cm} real training images & 15.22\% \\
      & 30,000 \hspace{0.001cm} real training images & 12.10\% \\
      & 68,120 \hspace{0.001cm} real training images & 7.54\% \\
	\hline
	\end{tabular}
\vspace{0.1cm}
\caption{Quantitative results on the Street View House Number (SVHN) format1 dataset~\cite{netzer2011reading}.}
\label{tab:SVHN}
\end{table}

\section{Conclusions}
\label{sec:conclusions}
Collecting a large scale supervised histopathology image dataset is extremely time consuming. We presented a complete pipeline for synthesizing realistic histopathology images with nucleus segmentation masks, which can be used for training supervised methods. Our method synthesizes images in various styles, utilizing textures and colors in real images. We train a task-specific CNN and a Generative Adversarial Network (GAN) in an end-to-end fashion, so that we can synthesize challenging training examples for the task-specific CNN on-the-fly. We evaluate our approach on the nucleus segmentation task. When no supervised data exists for a cancer type, our result is significantly better than across-cancer generalization results by supervised methods. Additionally, even when supervised data exists, our approach performed better than supervised methods. In the future, We plan to incorporate additional supervised classification and segmentation methods in our framework. Furthermore, we plan to model the texture of nuclei more accurately in the initial synthesis phase.

\paragraph{Acknowledgements} This work was supported in part by 1U24CA180924-01A1 from the NCI, R01LM011119-01 and R01LM009239 from the NLM, the Stony Brook University SensorCAT, a gift from Adobe, and the Partner University Fund 4DVision project.

{\small
\bibliographystyle{ieee}
\bibliography{egbib}
}

\end{document}